\documentclass[11pt,a4paper]{article}
\usepackage[hyperref]{naaclhlt2018}
\usepackage{times}
\usepackage{latexsym}
\usepackage{url}

\usepackage{graphicx}
\usepackage{wrapfig}
\usepackage{amsmath}
\usepackage{array,multirow}
\usepackage{booktabs}
\usepackage{tikz}
\usepackage{color}

\usetikzlibrary{arrows}
\usetikzlibrary{arrows.meta}
\usetikzlibrary{calc}
\usetikzlibrary{shapes}

\aclfinalcopy 

\title{A Deep Ensemble Model with Slot Alignment for Sequence-to-Sequence Natural Language Generation}
  
\author{Juraj Juraska, Panagiotis Karagiannis, Kevin K. Bowden \and Marilyn A. Walker \\
Natural Language and Dialogue Systems Lab \\
University of California, Santa Cruz \\
{\tt \{jjuraska,pkaragia,kkbowden,mawalker\}@ucsc.edu }}

\date{}

\begin{document}

\maketitle

\begin{abstract}

Natural language generation lies at the core of generative dialogue systems and conversational agents. We describe an ensemble neural language generator, and present several novel methods for data representation and augmentation that yield improved results in our model. We test the model on three datasets in the restaurant, TV and laptop domains, and report both objective and subjective evaluations of our best model. Using a range of automatic metrics, as well as human evaluators, we show that our approach achieves better results than state-of-the-art models on the same datasets.

\end{abstract}

\section{Introduction}
\label{sec:introduction}

There has recently been a substantial amount of research in natural language processing (NLP) in the context of  personal assistants, such as Cortana or Alexa. The capabilities of these conversational agents are still fairly limited and lacking in various aspects, one of the most challenging of which is the ability to produce utterances with human-like coherence and naturalness for many different kinds of content. This is the responsibility of the natural language generation (NLG) component.

Our work focuses on language generators whose inputs are  structured \emph{meaning representations}~(MRs). An MR describes a single dialogue act with a list of key concepts which need to be conveyed to the human user during the dialogue. Each piece of information is represented by a slot-value pair, where the \emph{slot} identifies the type of information and the \emph{value} is the corresponding content. \emph{Dialogue act}~(DA) types vary depending on the dialogue manager, ranging from simple ones, such as a \emph{goodbye} DA with no slots at all, to complex ones, such as an \emph{inform} DA containing multiple slots with various types of values (see example in Table~\ref{table:mr_utterance_example}).

\begin{table}
    \small
   	\centering
    \begin{tabular}{>{\centering\arraybackslash} m{0.1\linewidth} m{0.79\linewidth}}
    	\toprule
    	\textbf{MR} & \emph{inform} (name [\textbf{The Golden Curry}], food [\textbf{Japanese}], priceRange [\textbf{moderate}], familyFriendly [\textbf{yes}], near [\textbf{The Bakers}]) \\
        \midrule
    	\textbf{Utt.} & Located near \textbf{The Bakers}, \textbf{kid-friendly} restaurant, \textbf{The Golden Curry}, offers \textbf{Japanese} cuisine with a \textbf{moderate} price range. \\
        \bottomrule
    \end{tabular}
 	\vspace{-0.3cm}
	\caption{An example of an MR and a corresponding reference utterance.}
    \label{table:mr_utterance_example}
 	\vspace{-0.3cm}
\end{table}


A natural language generator must produce a syntactically and semantically correct utterance from a given MR. The utterance should express all the information contained in the MR, in a natural and conversational way. In traditional language generator architectures, the assembling of an utterance from an MR is performed in two stages: \emph{sentence planning}, which enforces semantic correctness and determines the structure of the utterance, and \emph{surface realization}, which enforces syntactic correctness and produces the final utterance form.

Earlier work on statistical NLG approaches were typically hybrids of a handcrafted component and a statistical training method \cite{langkilde1998generation, stent2004trainable, rieser2010natural}.
The handcrafted aspects, however, lead to decreased portability and potentially limit the variability of the outputs. New corpus-based approaches emerged that used semantically aligned data to train language models that output utterances directly from their MRs \cite{mairesse2010phrase, mairesse2014stochastic}. The alignment provides valuable information during training, but the semantic annotation is costly.

The most recent methods do not require aligned data and use an \emph{end-to-end} approach to training, performing sentence planning and surface realization simultaneously \cite{konstas2013global}. The most successful systems trained on unaligned data use recurrent neural networks (RNNs) paired with an encoder-decoder system design \cite{mei2015talk, duvsek2016sequence}, but also other concepts, such as imitation learning \cite{lampouras2016imitation}. These NLG models, however, typically require greater amount of data for training due to the lack of semantic alignment, and they still have problems producing syntactically and semantically correct output, as well as being limited in naturalness \cite{nayak2017plan}.

Here we present a neural ensemble natural language generator, which we train and test on three large unaligned datasets in the restaurant, television, and laptop domains. We explore novel ways to represent the MR inputs, including novel methods for delexicalizing slots and their values, automatic slot alignment, as well as the use of a semantic reranker. We use automatic evaluation metrics to show that these methods appreciably improve the performance of our model. On the largest of the datasets, the E2E dataset \cite{novikova2017e2e} with nearly 50K samples, we also demonstrate that our model significantly outperforms the baseline \emph{E2E NLG Challenge}\footnote{http://www.macs.hw.ac.uk/InteractionLab/E2E/} system in human evaluation. Finally, after augmenting our model with stylistic data selection, subjective evaluations reveal that it can still produce overall better results despite a significantly reduced training set.


\section{Related Work}
\label{sec:related_work}

NLG is closely related to machine translation and has similarly benefited from recent rapid development of deep learning methods. State-of-the-art NLG systems build thus on deep neural \emph{sequence-to-sequence} models \cite{sutskever2014sequence} with an \emph{encoder-decoder} architecture \cite{cho2014learning} equipped with an \emph{attention} mechanism \cite{bahdanau2015neural}. They typically also rely on slot \emph{delexicalization} \cite{mairesse2010phrase, henderson2014robust}, which allows the model to better generalize to unseen inputs, as exemplified by TGen \cite{duvsek2016sequence}. However, \citet{nayak2017plan} point out that there are frequent scenarios where delexicalization behaves inadequately (see Section~\ref{subsec:delex} for more details), and \citet{agarwal2017surprisingly} show that a \emph{character-level} approach to NLG may avoid the need for delexicalization, at the potential cost of making more semantic omission errors.

The end-to-end approach to NLG typically requires a mechanism for aligning slots on the output utterances: this allows the model to generate utterances with fewer missing or redundant slots. \citet{cuayahuitl2014training} perform automatic slot labeling using a Bayesian network trained on a labeled dataset, and show that a method using spectral clustering can be extended to unlabeled data with high accuracy. In one of the first successful neural approaches to language generation, \citet{wen2015stochastic} augment the generator's inputs with a control vector indicating which slots still need to be realized at each step. \citet{wen2015semantically} take the idea further by embedding a new sigmoid gate into their LSTM cells, which directly conditions the generator on the DA. More recently, \citet{duvsek2016sequence} supplement their encoder-decoder model with a trainable classifier which they use to rerank the beam search candidates based on missing and redundant slot mentions.

Our work builds upon the successful attentional encoder-decoder framework for sequence-to-sequence learning and expands it through ensembling. We explore the feasibility of a domain-independent slot aligner that could be applied to any dataset, regardless of its size, and beyond the reranking task. We also tackle some challenges caused by delexicalization in order to improve the quality of surface realizations, while retaining the ability of the neural model to generalize.

\section{Datasets}
\label{sec:datasets}

We evaluated the models on three datasets from different domains. The primary one is the recently released E2E restaurant dataset~\cite{novikova2017e2e} with 48K samples. For benchmarking we use the TV dataset and the Laptop dataset~\cite{wen2016multi} with 7K and 13K samples, respectively. Table~\ref{table:dataset_overview} summarizes the proportions of the training, validation, and test sets for each dataset.

\begin{table}
  \small
  \centering
  \begin{tabular}{c r r r}
    \toprule
    & \textbf{E2E}	& \textbf{TV}	& \textbf{Laptop} \\
    \midrule
    $|$training set$|$		& 42061	& 4221	& 7944 \\
    $|$validation set$|$	& 4672	& 1407	& 2649 \\
    $|$test set$|$			& 630	& 1407	& 2649 \\
    total					& 47363	& 7035	& 13242 \\
    \midrule
    DA types	& 1		& 14	& 14 \\
    slot types	& 8		& 16	& 20 \\
    \bottomrule
  \end{tabular}
  \vspace{-0.1cm}
  \caption{Overview of the number of samples, as well as different DA and slot types, in each dataset
  \label{table:dataset_overview}.}
  \vspace{-0.2cm}
\end{table}

\begin{figure}
  \begin{center}
  	\includegraphics[width=\columnwidth]{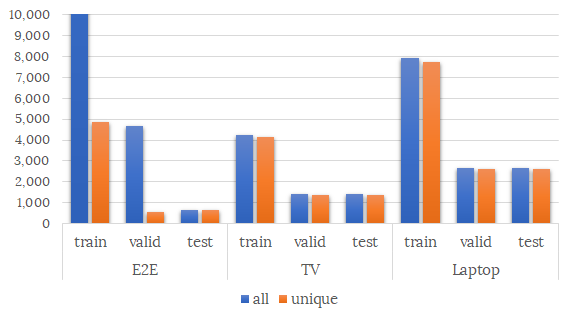}
  \end{center}
  \vspace{-0.5cm}
  \caption{Proportion of unique MRs in the datasets. Note that the number of MRs in the E2E dataset was cut off at 10K for the sake of visibility of the small differences between other column pairs.}
  \label{fig:unique_mrs}
  \vspace{-0.2cm}
\end{figure}

\subsection{E2E Dataset}

The E2E dataset is by far the largest one available for task-oriented language generation in the restaurant domain. The human references were collected using pictures as the source of information, which was shown to inspire more informative and natural utterances~\cite{novikova2016crowd}. With nearly 50K samples, it offers almost 10 times more data than the San Francisco restaurant dataset introduced in~\citet{wen2015semantically}, which has frequently been used for benchmarks. The reference utterances in the E2E dataset exhibit superior lexical richness and syntactic variation, including more complex discourse phenomena. It aims to provide higher-quality training data for end-to-end NLG systems to learn to produce more naturally sounding utterances. The dataset was released as a part of the E2E NLG Challenge.

Although the E2E dataset contains a large number of samples, each MR is associated on average with $8.65$ different reference utterances, effectively offering less than 5K unique MRs in the training set (Fig.~\ref{fig:unique_mrs}). Explicitly providing the model with multiple ground truths, it offers multiple alternative utterance structures the model can learn to apply for the same type of MR. The delexicalization, as detailed later in Section~\ref{subsec:delex}, improves the ability of the model to share the concepts across different MRs.

The dataset contains only 8 different slot types, which are fairly equally distributed. The number of slots in each MR ranges between 3 and 8, but the majority of MRs consist of 5 or 6 slots. Even though most of the MRs contain many slots, the majority of the corresponding human utterances, however, consist of one or two sentences only (Table~\ref{table:sentence_average}), suggesting a reasonably high level of sentence complexity in the references.

\begin{table}
  \small
  \centering
  \begin{tabular}{l r r r r r r}
    \toprule
    \textbf{slots}	& 3	& 4	& 5	& 6	& 7	& 8 \\
    \midrule
    \textbf{sent.}	& 1.09 & 1.23 & 1.41 & 1.65 & 1.84 & 1.92 \\
    \textbf{prop.}	& 5\% & 18\% & 32\% & 28\% & 14\% & 3\% \\
    \bottomrule
  \end{tabular}
  \caption{Average number of sentences in the reference utterance for a given number of slots in the corresponding MR, along with the proportion of MRs with specific slot counts.}
  \label{table:sentence_average}
\end{table}


\subsection{TV and Laptop Datasets}

The reference utterances in the TV and the Laptop datasets were collected using Amazon Mechanical Turk (AMT), one utterance per MR. These two datasets are similar in structure, both using the same 14 DA types.\footnote{We noticed the MRs with the \texttt{?request} DA type in the TV dataset have no slots provided, as opposed to the Laptop dataset, so we imputed these in order to obtain valid MRs.} The Laptop dataset, however, is almost twice as large and contains 25\% more slot types.

\begin{figure}
  \begin{center}
  	\includegraphics[width=\columnwidth]{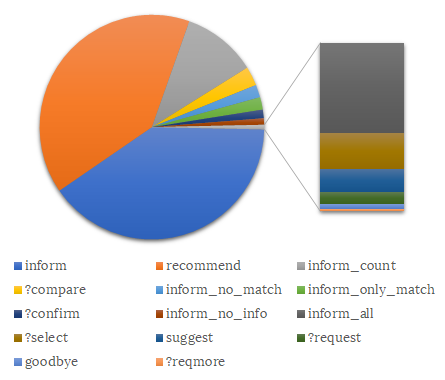}
  \end{center}
  \vspace{-0.5cm}
  \caption{Proportion of DAs in the Laptop dataset.}
  \label{fig:da_freq_laptop}
\end{figure}

Although both of these datasets contain more than a dozen different DA types, the vast majority (68\% and 80\% respectively) of the MRs describe a DA of either type \texttt{inform} or \texttt{recommend} (Fig.~\ref{fig:da_freq_laptop}), which in most cases have very similarly structured realizations, comparable to those in the E2E dataset. DAs such as \texttt{suggest}, \texttt{?request}, or \texttt{goodbye} are represented by less than a dozen samples, but are significantly easier to learn to generate an utterance from because the corresponding MRs contain three slots at the most. 

\section{Ensemble Neural Language Generator}
\label{model}

\subsection{Encoder-Decoder with Attention}

Our model uses the standard encoder-decoder architecture with attention, as defined in \citet{bahdanau2015neural}. Encoding the input into a sequence of context vectors instead of a single vector enables the decoder to learn what specific parts of the input sequence to pay attention to, given the output generated so far. In this attentional encoder-decoder architecture, the probability of the output at each time step $t$ of the decoder depends on a distinct context vector $q_t$ in the following way:
\[ P(u_t | u_1,\ldots, u_{t-1}, \mathbf{w}) = g(u_{t-1}, s_t, q_t) \;, \]
where in the place of function $g$ we use the softmax function over the size of the vocabulary, and $s_t$ is a hidden state of the decoder RNN at time step $t$, calculated as:
\[ s_t = f(s_{t-1}, u_{t-1}, q_t) \;. \]
The context vector $q_t$ is obtained as a weighted sum of all the hidden states $h_1,\ldots, h_L$ of the encoder:
\[ q_t = \sum_{i=1}^{L} \alpha_{t,i} h_i \;, \]
where $\alpha_{t,i}$ corresponds to the attention score the $t$-th word in the target sentence assigns to the $i$-th item in the input MR.

\begin{figure}
  \begin{center}
  	\includegraphics[width=\columnwidth]{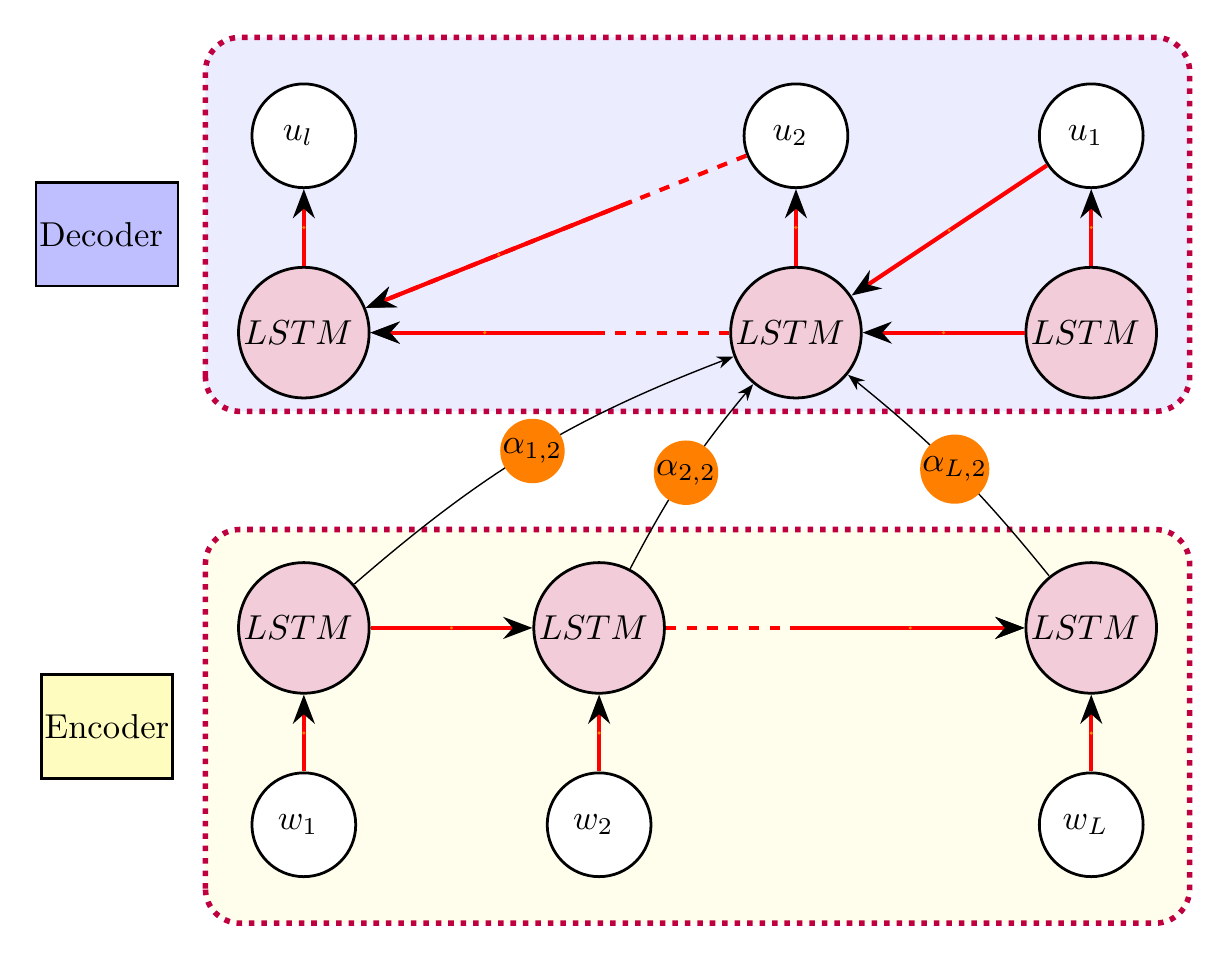}
  \end{center}
  \vspace{-0.4cm}
  \caption{Standard architecture of a single-layer encoder-decoder LSTM model with attention. For each time step $t$ in the output sequence, the attention scores $\alpha_{t,1}, \dots, \alpha_{t,L}$ are calculated. This diagram shows the attention scores only for $t=2$.}
  \label{fig:enc_dec}
\end{figure}

We compute the attention score $\alpha_{t,i}$ using a multi-layer perceptron~(MLP) jointly trained with the entire system \cite{bahdanau2015neural}. The encoder's and decoder's hidden states at time $i$ and~$t$, respectively, are concatenated and used as the input to the MLP, namely:
\[ \alpha_{t,i} = softmax\left( \mathbf{w}^T tanh\left(W[h_i ; s_t]\right) \right) \;, \] 
where $W$ and $\mathbf{w}$ are the weight matrix and the vector of the first and the second layer of the MLP, respectively. The learned weights indicate the level of influence of the individual words in the input sequence on the prediction of the word at time step $t$ of the decoder. The model thus learns a soft alignment between the source and the target sequence.

\subsection{Ensembling}
\label{sec:ensembling}

In order to enhance the quality of the predicted utterances, we create three neural models with different encoders. Two of the models use a bidirectional LSTM \cite{hochreiter1997long} encoder, whereas the third model has a CNN \cite{lecun1998gradient} encoder. We train these models individually for a different number of epochs and then combine their predictions.

Initially, we attempted to combine the predictions of the models by averaging the log-probability at each time step and then selecting the word with the maximum log-probability. We noticed that the quality, as well as the BLEU score of our utterances, decreased significantly. We believe that this is due to the fact that different models learn different sentence structures and, hence, combining predictions at the probability level results in incoherent utterances.

Therefore, instead of combining the models at the log-probability level, we accumulate the top~10 predicted utterances from each model type using beam search and allow the reranker (see Section~\ref{subsec:reranker}) to rank all candidate utterances taking the proportion of slots they successfully realized into consideration. Finally, our system predicts the utterance that received the highest score.

\subsection{Slot Alignment}
\label{subsec:slot_alignment}

Our training data is inherently unaligned, meaning our model is not certain which sentence in a multi-sentence utterance contains a given slot, which limits the model's robustness. To accommodate this, we create a heuristic-based slot aligner which automatically preprocesses the data. Its primary goal is to align chunks of text from the reference utterances with an expected value from the MR. Applications of our slot aligner are described in subsequent sections and in Table \ref{table:utt_splitting_example}.

In our task, we have a finite set of slot mentions which must be detected in the corresponding utterance. Moreover, from our training data we can see that most slots are realized by inserting a specific set of phrases into an utterance. Using this insight, we construct a gazetteer, which primarily searches for overlapping content between the MR and each sentence in an utterance, by associating all possible slot realizations with their appropriate slot type. We additionally augment the gazetteer using a small set of handcrafted rules which capture cases not easily encapsulated by the above process, for example, associating the \texttt{priceRange} slot with a chunk of text using currency symbols or relevant lexemes, such as ``cheap'' or ``high-end''. While handcrafted, these rules are transferable across domains, as they target the slots, not the domains, and mostly serve to counteract the noise in the E2E dataset. Finally, we use WordNet \cite{fellbaum1998wordnet} to further augment the size of our gazetteer by accounting for synonyms and other semantic relationships, such as associating ``pasta'' with the \texttt{food[Italian]} slot.

\subsection{Reranker}
\label{subsec:reranker}

As discussed in Section \ref{sec:ensembling}, our model uses beam search to produce a pool of the most likely utterances for a given MR. While these results have a probability score provided by the model, we found that relying entirely on this score often results in the system picking a candidate which is objectively worse than a lower scoring utterance (i.e. one missing more slots and/or realizing slots incorrectly). We therefore augment that score by multiplying it by the following score which takes the slot alignment into consideration:
\[ s_\text{align} = \frac{N}{(N_u + 1) \cdot (N_o + 1)} \;, \]
where $N$ is the number of all slots in the given MR, and $N_u$ and $N_o$ represent the number of unaligned slots (those not observed by our slot aligner) and over-generated slots (those which have been realized but were not present in the original MR), respectively.

\section{Data Preprocessing}
\label{sec:data}

\subsection{Delexicalization}
\label{subsec:delex}

We enhance the ability of our model to generalize the learned concepts to unseen MRs by delexicalizing the training data. Moreover, it reduces the amount of data required to train the model. We identify the categorical slots whose values always propagate verbatim to the utterance, and replace the corresponding values in the utterance with placeholder tokens. The placeholders are eventually replaced in the output utterance in post-processing  by copying the  values from the input MR. Examples of such slots would be \texttt{name} or \texttt{near} in the E2E dataset, and \texttt{screensize} or \texttt{processor} in the TV and the Laptop dataset.

Previous work identifies categorical slots as good delexicalization candidates that improve  the performance of the model \cite{wen2015semantically,nayak2017plan}. However, we chose not to delexicalize those categorical slots whose values can be expressed in alternative ways, such as ``less than \$20'' and ``cheap'',  or ``on the riverside'' and ``by the river''. Excluding these from delexicalization may lead to an increased number of incorrect realizations, but it encourages diversity of the model's outputs by giving it a freedom to choose among alternative ways of expressing a slot-value in different contexts. This, however, assumes that the training set contains a sufficient number of samples displaying this type of alternation so that the model can learn that certain phrases are synonymous. With its multiple human references for each MR, the E2E dataset has this property.

As \citet{nayak2017plan} point out, delexicalization affects the sentence planning and the lexical choice around the delexicalized slot value. For example, the realization of the slot \texttt{food[\emph{Italian}]} in the phrase ``serves \emph{Italian} food'' is valid, while the realization of \texttt{food[\emph{fast food}]} in ``serves \emph{fast food} food'' is clearly undesired. Similarly, a naive delexicalization can result in ``a Italian restaurant'', whereas the article should be ``an''.  Another problem with articles is singular versus plural nouns in the slot value. For example, the slot \texttt{accessories} in the TV dataset, can take on values such as ``remote control'', as well as ``3D glasses'', where only the former  requires an article before the value.

We tackle this issue by defining different placeholder tokens for values requiring different treatment in the realization. For instance, the value ``Italian'' of the \texttt{food} slot is replaced by \texttt{slot\_vow\_cuisine\_food}, indicating that the value starts with a vowel and represents a cuisine, while ``fast food'' is replaced by \texttt{slot\_con\_food}, indicating that the value starts with a consonant and cannot be used as a term for cuisine. The model thus learns to generate ``a'' before \texttt{slot\_con\_food} and ``an'' before \texttt{slot\_vow\_cuisine\_food} when appropriate, as well as to avoid generating the word ``food'' after \texttt{food}-slot placeholders that do not contain the word ``cuisine''. All these rules are general and can automatically be applied across different slots and domains.

\subsection{Data Expansion}

\subsubsection*{Slot Permutation}

In our initial experiments, we tried expanding the training set by permuting the slot ordering in the MRs as suggested in~\citet{nayak2017plan}. From different slot orderings of every MR we sampled five random permutations (in addition to the original MR), and created new pseudo-samples with the same reference utterance. The training set thus increased six times in size.

Using such an augmented training set might add to the model's robustness, nevertheless it did not prove to be helpful with the E2E dataset. In this dataset, we observed the slot order to be fixed across all the MRs, both in the training and the test set. As a result, for the majority of the time, the model was training on MRs with slot orders it would never encounter in the test set, which ultimately led to a decreased performance in prediction on the test set.

\subsubsection*{Utterance/MR Splitting}

Taking a more utterance-oriented approach, we augment the training set with single-sentence utterances paired with their corresponding MRs. These new pseudo-samples are generated by splitting the existing reference utterances into single sentences and using the slot aligner introduced in Section~\ref{subsec:slot_alignment} to identify the slots that correspond to each sentence. The MRs of the new samples are created as the corresponding subsets of slots and, whenever the sentence contains the name (of the restaurant/TV/etc.) or a pronoun referring to it (such as ``it'' or ``its''), the \texttt{name} slot is included too. Finally, a new \texttt{position} slot is appended to every new MR, indicating whether it represents the first sentence or a subsequent sentence in the original utterance. An example of this splitting technique can be seen in Table~\ref{table:utt_splitting_example}. The training set almost doubled in size through this process.

\begin{table}
    \small
   	\centering
    \begin{tabular}{>{\centering\arraybackslash} m{0.12\linewidth} m{0.77\linewidth} }
    	\toprule
    	\textbf{MR} & \textcolor{magenta}{name [The Waterman],} \textcolor{red}{food [English],} \textcolor{blue}{priceRange [cheap],} \textcolor{red}{customer rating [average],} \textcolor{blue}{area [city centre], familyFriendly [yes]} \\
        \midrule
        \textbf{Utt.} & \textcolor{blue}{There is a family-friendly, cheap restaurant in the city centre, called The Waterman.} \textcolor{red}{It serves English food and has an average rating by customers.} \\
        \midrule
        \textbf{New MR \#1} & \textcolor{blue}{name [The Waterman], priceRange [cheap], area [city centre], familyFriendly [yes]}, \emph{position [outer]} \\
        \midrule
        \textbf{New MR \#2} & \textcolor{red}{name [The Waterman], food [English], customer rating [average]}, \emph{position [inner]} \\
        \bottomrule
    \end{tabular}
  	\vspace{-0.3cm}
	\caption{An example of the utterance/MR splitting.}
    \label{table:utt_splitting_example}
  	\vspace{0.1cm}
\end{table}

Since the slot aligner works heuristically, not all utterances are successfully aligned with the MR. The vast majority of such cases, however, is caused by reference utterances in the datasets having incorrect or entirely missing slot mentions. There is a noticeable proportion of those, so we leave them in the training set with the unaligned slots removed from the MR so as to avoid confusing the model when learning from such samples.


\subsection{Sentence Planning via Data Selection}
\label{subsec:data_filtering}

The quality of the training data inherently imposes an upper bound on the quality of the predictions of our model. Therefore, in order to bring our model to produce more sophisticated utterances, we experimented with filtering the training data to contain only the most natural sounding and structurally complex utterances for each MR. For instance, we prefer having an elegant, single-sentence utterance with an apposition as the reference for an MR, rather than an utterance composed of three simple sentences, two of which begin with ``it'' (see the examples in Table~\ref{table:utterance_style_examples}).

\begin{table}
    \small
   	\centering
    \begin{tabular}{>{\centering\arraybackslash} m{0.15\linewidth} m{0.74\linewidth}}
    	\toprule
    	\textbf{MR} & name [Wildwood], eatType [coffee shop], food [English], priceRange [moderate], customer rating [1 out of 5], near [Ranch] \\
        \midrule
    	\textbf{Simple utt.} & Wildwood provides English food for a moderate price. It has a low customer rating and is located near Ranch. It is a coffee shop. \\
        \midrule
    	\textbf{Elegant utt.} & A low-rated English style coffee shop around Ranch, called Wildwood, has moderately priced food. \\
        \bottomrule
    \end{tabular}
 	\vspace{-0.2cm}
	\caption{Contrastive example of a simple and a more elegant reference utterance style for the same MR in the E2E dataset.}
    \label{table:utterance_style_examples}
\end{table}

We assess the complexity and naturalness of each utterance by the use of discourse phenomena, such as contrastive cues, subordinate clauses, or aggregation. We identify these in the utterance's parse-tree produced by the Stanford CoreNLP toolkit \cite{manning2014stanford} by defining a set of rules for extracting the discourse phenomena. Furthermore, we consider the number of sentences used to convey all the information in the corresponding MR, as longer sentences tend to exhibit more advanced discourse phenomena. Penalizing utterances for too many sentences contributes to reducing the proportion of generic reference utterances, such as the ``simple'' example in the above table, in the filtered training set.

\section{Evaluation}
\label{sec:evaluation}

Researchers in NLG have generally used both automatic and human evaluation. Our results report the standard automatic evaluation metrics: BLEU \cite{papineni2002bleu}, NIST \cite{przybocki2009nist}, METEOR \cite{lavie2007meteor}, and ROUGE-L \cite{lin2004rouge}. For the E2E dataset experiments, we additionally report the results of the human evaluation carried out on the CrowdFlower platform as a part of the E2E NLG Challenge.

\subsection{Experimental Setup}

We built our ensemble model using the seq2seq framework~\cite{britz2017massive} for TensorFlow. Our individual LSTM models use a bidirectional LSTM encoder with 512 cells per layer, and the CNN models use a pooling encoder as in~\citet{gehring2017convolutional}. The decoder in all models was a 4-layer RNN decoder with 512 LSTM cells per layer and with attention. The hyperparameters were determined empirically. After experimenting with different beam search parameters, we settled on the beam width of 10. Moreover, we employed the length normalization of the beams as defined in~\citet{wu2016google}, in order to encourage the decoder to favor longer sequences. The length penalty providing the best results on the E2E dataset was 0.6, whereas for the TV and Laptop datasets it was 0.9 and 1.0, respectively.


\subsection{Experiments on the E2E Dataset}

We start by evaluating our system on the E2E dataset. Since the reference utterances in the test set were kept secret for the E2E NLG Challenge, we carried out the metric evaluation using the validation set. This was necessary to narrow down the models that perform well compared to the baseline. The final model selection was done based on a human evaluation of the models' outputs on the test set.

\subsubsection{Automatic Metric Evaluation}

In the first experiment, we assess what effect the augmenting of the training set via utterance splitting has on the performance of different models. The results in Table~\ref{table:results_utt_splitting} show that both the LSTM and the CNN models clearly benefit from additional pseudo-samples in the training set. This can likely be attributed to the model having access to more granular information about which parts of the utterance correspond to which slots in the MR. This may assist the model in sentence planning and building a stronger association between parts of the utterance and certain slots, such as that ``it'' is a substitute for the name.

\begin{table}
  \centering
  \begin{tabular} {m{0.8cm} >{\centering\arraybackslash}m{0.1cm} >{\centering\arraybackslash}m{1cm} >{\centering\arraybackslash}m{0.7cm} >{\centering\arraybackslash}m{1.4cm} >{\centering\arraybackslash}m{1.2cm} }
    \toprule
    &
    & \textbf{BLEU}
    & \textbf{NIST} 
    & \textbf{METEOR}
    & \textbf{ROUGE} \\
    \midrule
    \multirow{2}{*}{\textbf{LSTM}} 	& $\overline{\mathbf{s}}$	& 0.6664	& 8.0150 &0.4420 &0.7062 \\
    	& $\mathbf{s}$		& 0.6930$^\ddagger$	& 8.4198	& 0.4379	& 0.7099 \\
    \midrule
    \multirow{2}{*}{\textbf{CNN}} 	& $\overline{\mathbf{s}}$	& 0.6599	& 7.8520	& 0.4333	& 0.7018 \\
    	& $\mathbf{s}$		& 0.6760$^\dagger$	& 8.0440	& 0.4448	& 0.7055 \\
    \bottomrule
  \end{tabular}
  \vspace{-0.1cm}
  \caption{Automatic metric scores of different models tested on the E2E dataset, both unmodified ($\overline{\mathbf{s}}$) and augmented ($\mathbf{s}$) through the utterance splitting. The symbols $^\dagger$ and $^\ddagger$ indicate statistically significant improvement over the $\overline{\mathbf{s}}$ counterpart with $p < 0.05$ and $p < 0.01$, respectively, based on the paired t-test.}
  \label{table:results_utt_splitting}
  \vspace{-0.1cm}
\end{table}

Testing our ensembling approach reveals that reranking predictions pooled from different models produces an ensemble model that is overall more robust than the individual submodels. The submodels fail to perform well in all four metrics at once, whereas the ensembling creates a new model that is more consistent across the different metric types (Table~\ref{table:results_ensemble_devset}).\footnote{The scores here correspond to the model submitted to the E2E NLG Challenge. Subsequently, we found better performing models according to some metrics: see Table~\ref{table:results_utt_splitting}.} While the ensemble model decreases the proportion of incorrectly realized slots compared to its individual submodels on the validation set, on the test set it only outperforms two of the submodels in this aspect (Table~\ref{table:results_ensemble_err}). Analyzing the outputs, we also observed that the CNN model surpassed the two LSTM models in the ability to realize the ``fast food'' and ``pub'' values reliably, both of which were hardly present in the validation set but very frequent in the test set. On the official E2E test set, our ensemble model performs comparably to the baseline model, TGen~\cite{duvsek2016sequence}, in terms of automatic metrics (Table~\ref{table:results_ensemble_testset}).

\begin{table}
  \centering
  \begin{tabular}{m{1.1cm} >{\centering\arraybackslash}m{0.9cm} >{\centering\arraybackslash}m{0.8cm} >{\centering\arraybackslash}m{1.5cm} >{\centering\arraybackslash}m{1.3cm}}
    \toprule
	& \textbf{BLEU}		
	& \textbf{NIST}
	& \textbf{METEOR}
	& \textbf{ROUGE} \\
    \midrule
    \textbf{LSTM1}	& 0.6661	& 8.1626	& 0.4644	& 0.7018 \\
    \textbf{LSTM2}	& 0.6493	& 7.9996	& 0.4649	& 0.6995 \\
    \textbf{CNN}	& 0.6636	& 7.9617	& 0.4700	& 0.7107 \\
    \midrule
    \textbf{Ensem.}	& 0.6576	& 8.0761	& 0.4675	& 0.7029 \\
    \bottomrule
  \end{tabular}
  \vspace{-0.2cm}
  \caption{Automatic metric scores of three different models and their ensemble, tested on the \emph{validation set} of the E2E dataset. LSTM2 differs from LSTM1 in that it was trained longer.}
  \label{table:results_ensemble_devset}
  \vspace{0.2cm}
\end{table}

\begin{table}
  \centering
  \begin{tabular}{l r r}
    \toprule
	& \textbf{Validation set}	& \textbf{Test set} \\
    \midrule
    \textbf{LSTM1}	& 0.116\%	& 0.988\% \\
    \textbf{LSTM2}	& 0.145\%	& 1.241\% \\
    \textbf{CNN}	& 0.232\%	& 0.253\% \\
    \midrule
    \textbf{Ensem.}	& 0.087\%	& 0.965\% \\
    \bottomrule
  \end{tabular}
  \caption{Error rate of the ensemble model compared to its individual submodels.}
  \label{table:results_ensemble_err}
\end{table}

\subsubsection{Human Evaluation}

It is known that automatic metrics function only as a general and vague indication of the quality of an utterance in a dialogue \cite{liu2016not, novikova2017we}. Systems which score similarly according to these metrics could produce utterances that are significantly different because automatic metrics fail to capture many of the characteristics of natural sounding utterances. Therefore, to better assess the structural complexity of the predictions of our model, we present the results of a human evaluation of the models' outputs in terms of both naturalness and quality, carried out by the E2E NLG Challenge organizers.

\emph{Quality} examines the grammatical correctness and adequacy of an utterance given an MR, whereas \emph{naturalness} assesses whether a predicted utterance could have been produced by a native speaker, irrespective of the MR. To obtain these scores, crowd workers ranked the outputs of 5 randomly selected systems from worst to best. The final scores were produced using the TrueSkill algorithm \cite{sakaguchi2014efficient} through pairwise comparisons of the human evaluation scores among the 20 competing systems.

Our system, trained on the E2E dataset without stylistic selection (Section~\ref{subsec:data_filtering}), achieved the highest quality score in the E2E NLG Challenge, and was ranked second in naturalness.\footnote{The system that surpassed ours in naturalness was ranked the last according to the quality metric.} The system's performance in quality (the primary metric) was significantly better than the competition according to the TrueSkill evaluation, which used bootstrap resampling with a $p$-level of $p \leq 0.05$. Comparing these results with the scores achieved by the baseline model in quality and naturalness (5th and 6th place, respectively) reinforces our belief that models that perform similarly on the automatic metrics (Table~\ref{table:results_ensemble_testset}) can exhibit vast differences in the structural complexity of their generated utterances.

\begin{table}
  \centering
  \begin{tabular}{m{1.1cm} >{\centering\arraybackslash}m{0.9cm} >{\centering\arraybackslash}m{0.8cm} >{\centering\arraybackslash}m{1.5cm} >{\centering\arraybackslash}m{1.3cm}}
    \toprule
	& \textbf{BLEU}		
	& \textbf{NIST}
	& \textbf{METEOR}
	& \textbf{ROUGE} \\
    \midrule
    \textbf{TGen}	& 0.6593	& 8.6094	& \emph{0.4483}	& \emph{0.6850} \\
    \midrule
    \textbf{Ensem.}	& \emph{0.6619}	& \emph{8.6130}	& 0.4454	& 0.6772 \\
    \bottomrule
  \end{tabular}
  \vspace{-0.2cm}
  \caption{Automatic metric scores of our ensemble model compared against TGen (the baseline model), tested on the \emph{test set} of the E2E dataset.}
  \label{table:results_ensemble_testset}
  \vspace{0.2cm}
\end{table}

\subsubsection{Experiments with Data Selection}

After filtering the E2E training set as described in Section~\ref{subsec:data_filtering}, the new training set consisted of approximately 20K pairs of MRs and utterances. Interestingly, despite this drastic reduction in training samples, the model was able to learn more complex utterances that contained the natural variations of the human language. The generated utterances exhibited discourse phenomena such as contrastive cues (see Example \#1 in Table~\ref{table:stylistic_selection_examples}), as well as a more conversational style (Example \#2). Nevertheless, the model also failed to realize slots more frequently.

\begin{table}
    \small
   	\centering
    \begin{tabular}{>{\centering\arraybackslash} m{0.05\linewidth} m{0.84\linewidth}}
    	\toprule
    	\textbf{Ex. \#1} & The Cricketers is a cheap Chinese restaurant near All Bar One in the riverside area, \textbf{but it has an average customer rating and is not family friendly}. \\
        \midrule
    	\textbf{Ex. \#2} & \textbf{If you are looking for} a coffee shop near The Rice Boat, \textbf{try} Giraffe. \\
        \bottomrule
    \end{tabular}
 	\vspace{-0.2cm}
	\caption{Examples of generated utterances that contain more advanced discourse phenomena.}
    \label{table:stylistic_selection_examples}
\end{table}

In order to observe the effect of stylistic data selection, we conducted a human evaluation where we assessed the utterances based on \emph{error rate} and \emph{naturalness}. The error rate is calculated as the percentage of slots the model failed to realize divided by the total number of slots present among all samples. The annotators ranked samples of utterance triples -- corresponding to three different ensemble models -- by naturalness from 1 to 3 (3 being the most natural, with possible ties). The \emph{conservative} model combines three submodels all trained on the full training set, the \emph{progressive} one combines submodels solely trained on the filtered dataset, and finally, the \emph{hybrid} is an ensemble of three models only one of which is trained on the full training set, so as to serve as a fallback.

The impact of the reduction of the number of training samples becomes evident by looking at the score of the progressive model (Table~\ref{table:results_naturalness}), where this model trained solely on the reduced dataset had the highest error rate. We observe, however, that a hybrid ensemble model manages to perform the best in terms of the error rate, as well as the naturalness.

These results suggest that filtering the dataset through careful data selection can help to achieve better and more natural sounding utterances. It significantly improves the model's ability to produce more elegant utterances beyond the ``[name] is... It is/has...'' format, which is only too common in neural language generators in this domain.

\begin{table}
  \centering
  \begin{tabular}{p{2.8cm} c c}
    \toprule
    \textbf{Ensemble model}& \textbf{Error rate}	& \textbf{Naturalness}	\\[0.25ex] 
    \midrule
    Conservative	& \emph{0.40\%}	& 2.196	\\
    Progressive		& 1.60\%	& 2.118	\\
    \midrule
    Hybrid			& \emph{0.40\%}	& \emph{2.435}	\\
    \bottomrule
  \end{tabular}
  \vspace{-0.1cm}
  \caption{Average error rate and naturalness metrics obtained from six annotators for different ensemble models.}
  \label{table:results_naturalness}
\end{table}

\subsection{Experiments on TV and Laptop Datasets}

In order to provide a better frame of reference for the performance of our proposed model, we utilize the RNNLG benchmark toolkit\footnote{https://github.com/shawnwun/RNNLG} to evaluate our system on two additional, widely used datasets in NLG, and compare our results with those of a state-of-the-art model, SCLSTM \cite{wen2015semantically}. As Table~\ref{table:results_ensemble_tv_laptop} shows, our ensemble model performs competitively with the baseline on the TV dataset, and it outperforms it on the Laptop dataset by a wide margin. We believe the higher error rate of our model can be explained by the significantly less aggressive slot delexicalization than the one used in SCLSTM. That, however, gives our model a greater lexical freedom and, with it, the ability to produce more natural utterances.

The model trained on the Laptop dataset is also a prime example of how an ensemble model is capable of extracting the best learned concepts from each individual submodel. By combining their knowledge and compensating thus for each other's weaknesses, the ensemble model can achieve a lower error rate, as well as a better overall quality, than any of the submodels individually.

\begin{table}
  \centering
  \begin{tabular}{m{1.6cm} >{\centering\arraybackslash}m{1cm} >{\centering\arraybackslash}m{1cm} >{\centering\arraybackslash}m{1cm} >{\centering\arraybackslash}m{1cm}}
    \toprule
	& \multicolumn{2}{c}{\textbf{TV}}
    & \multicolumn{2}{c}{\textbf{Laptop}} \\
	& \textbf{BLEU}
	& \textbf{ERR}
	& \textbf{BLEU}
	& \textbf{ERR} \\
    \midrule
    \textbf{SCLSTM}	& 0.5265	& 2.31\%	& 0.5116	& 0.79\% \\
    \midrule
    \textbf{LSTM}	& 0.5012	& 3.86\%	& 0.5083	& 4.43\% \\
    \textbf{CNN}	& 0.5287	& 1.87\%	& 0.5231	& 2.25\% \\
    \midrule
    \textbf{Ensem.}	& 0.5226	& 1.67\%	& 0.5238	& 1.55\% \\
    \bottomrule
  \end{tabular}
  \vspace{-0.1cm}
  \caption{Automatic metric scores of our ensemble model evaluated on the test sets of the TV and Laptop datasets, and compared against SCLSTM. The ERR column indicates the slot error rate, as computed by the RNNLG toolkit (for our models calculated in post-processing).}
  \label{table:results_ensemble_tv_laptop}
\end{table}

\section{Conclusion and Future Work}
\label{sec:conclusion}

In this paper we presented our ensemble attentional encoder-decoder model for generating natural utterances from MRs. Moreover, we presented novel methods of representing the MRs to improve performance. Our results indicate that the proposed utterance splitting applied to the training set greatly improves the neural model's accuracy and ability to generalize. The ensembling method paired with the reranking based on slot alignment also contributed to the increase in quality of the generated utterances, while minimizing the number of slots that are not realized during the generation. This also enables the use of a less aggressive delexicalization, which in turn stimulates diversity in the produced utterances.

We showed that automatic slot alignment can be utilized for expanding the training data, as well as for utterance reranking. Our alignment currently relies in part on empirically observed heuristics, and a more robust aligner would allow for more flexible expansion into new domains. Since the stylistic data selection noticeably improved the diversity of our system's outputs, we believe this is a method with future potential, which we intend to further explore. Finally, it is clear that current automatic evaluation metrics in NLG are only sufficient for providing a vague idea as to the system's performance; we postulate that leveraging the reference data to train a classifier will result in a more conclusive automatic evaluation metric.

\section*{Acknowledgements}
This research was partially supported by NSF Robust Intelligence \#IIS-1302668-002.

\bibliographystyle{acl_natbib}
\bibliography{references}

\end{document}